# Comparing Active Learning Performance Driven by Gaussian Processes or Bayesian Neural Networks for Constrained Trajectory Exploration

Sapphira Akins[1], Frances Zhu[2]
*University of Hawaiʻi at Mānoa [1], Honolulu, HI, 96822, USA*

Robots with increasing autonomy progress our space exploration capabilities, particularly for in-situ exploration and sampling to stand in for human explorers. Currently, humans drive robots to meet scientific objectives, but depending on the robot's location, the exchange of information and driving commands between the human operator and robot may cause undue delays in mission fulfillment. An autonomous robot encoded with a scientific objective and an exploration strategy incurs no communication delays and can fulfill missions more quickly. Active learning algorithms offer this capability of intelligent exploration, but the underlying model structure varies the performance of the active learning algorithm in accurately forming an understanding of the environment. In this paper, we investigate the performance differences between active learning algorithms driven by Gaussian processes or Bayesian neural networks for exploration strategies encoded on agents that are constrained in their trajectories, like planetary surface rovers. These two active learning strategies were tested in a simulation environment against science-blind strategies to predict the spatial distribution of a variable of interest along multiple datasets. The performance metrics of interest are model accuracy in root mean squared (RMS) error, training time, model convergence, total distance traveled until convergence, and total samples until convergence. Active learning strategies encoded with Gaussian processes require less computation to train, converge to an accurate model more quickly, and propose trajectories of shorter distance, except in a few complex environments in which Bayesian neural networks achieve a more accurate model in the large data regime due to their more expressive functional bases. The paper concludes with advice on when and how to implement either exploration strategy for future space missions.

## I. Nomenclature

| | | |
|---|---|---|
| $d$ | = | distance |
| $d_c$ | = | distance until convergence |
| $D$ | = | dataset |
| $e$ | = | error |
| $f$ | = | true model |
| $\hat{f}$ | = | oracle model |
| $g$ | = | suggestion policy |
| $i$ | = | index |
| $i_c$ | = | samples until convergence |
| $J$ | = | objective function |
| $k$ | = | kernel |
| $N$ | = | normal distribution |
| $\mu$ | = | model posterior mean |
| $r$ | = | location in environment |
| $\mathcal{R}$ | = | the entire environment space |
| $\sigma$ | = | measurement noise |
| $t$ | = | time |
| $V$ | = | model posterior variance |
| $X$ | = | aggregate input dataset target position |
| $x$ | = | single training pair target position |
| $Y$ | = | aggregate input dataset target variable of interest |
| $y$ | = | single training pair target variable of interest |

## I. Introduction

Traditionally in robotic exploration either robots are teleoperated by humans or autonomous robots are provided with user-defined waypoints within the environment prior to deployment. There is always human involvement. Now, intelligent, adaptive autonomous robots are needed to explore unknown, dynamic environments where little is known a priori. The robot must use its own sensors to fully understand its environment. An in-situ exploration strategy that incorporates science information and maximizes a formal cost objective generating proximal destinations of interest, yielding more efficient scientific data collection, time savings, and potentially convergence properties. Even if this exploration strategy algorithm is not fully autonomous, the generated waypoints can inform teleoperators of potential destinations of interest, which could accelerate the site selection process or affirm sites selected by teleoperators.

The science mission that motivates this technology is the search for water ice. Water ice is one of the most important resources on the Moon and Mars [1], [2]. The direct detection of surface-exposed water ice using infrared data in the lunar polar regions accelerates the progress of exploring lunar ice in-situ resources [3]. Data gathered from observations of surface-level water-ice deposits on the Moon suggest these deposits may also exist subsurface. However, we do not currently have the knowledge necessary to classify any subset of the total volume of lunar water-ice resources. Orbital InfraRed (IR) measurements suggest that water ice exists in approximately 5% of Lunar cold traps (regions where the annual maximum temperature is less than 110 K and water-ice is stable) and in up to 30% of the total exposed surface mass [3]. At present, we do not yet understand enough about the physical characteristics of lunar water-ice deposits to consider these reserves for future exploration and resource utilization efforts. The most direct way to characterize the volume of subsurface water is to conduct an in-situ investigation, necessitating human or robot surface operations.

Currently, human operators intuit the scientific value of exploring specific destinations, much like NASA's Sojourner, China's Yutu-2, and MERs [4]. Although the most recently landed rover MSL shows hints of autonomy, the autonomous interactions are restricted to mobility actions – separate from any science [5]. Rover will very likely face power and thermal limitations dependent on time spent in a permanently shadowed region for which the mission cannot afford extensive sampling or teleoperators to stop and intuit the next waypoint to visit. The optimization problem of space exploration is that a limited set of spacecraft resources (power) must be allocated between competing choices (destinations) in a way that maximizes science discovered and mitigates risk, a specific formulation of the Bayesian optimization problem [6].

This paper directly compares the performance of active learning strategies driven by a Gaussian process or Bayesian neural network along metrics of accuracy (RMS error), train time, and samples until convergence in a constrained trajectory exploration application. Section II reviews core concepts in understanding Gaussian process performance to neural network performance in driving active learning algorithms and distinguishes this work from previous work. Section III discusses the active learning algorithm, the algorithm implementation, the benchmark environments, and the experiments run to compare Gaussian processes to Bayesian neural networks. Section IV reports the results of the comparison by defining the metrics for comparison, performance along these metrics, and an interpretation of performance for other applications.

## II. Background

Active learning algorithms use historical measurements to generate an uncertainty map that suggests a location in the space with the highest uncertainty to sample next, which offers a sample-efficient method for exploring and characterizing a space. The agent is encoded with an objective function, $J$, that aims to minimize a learned model's prediction $\hat{f}(X, t, D, k(\cdot))$ with respect to ground truth $f(X, t)$ at a location on the surface across a set of discretized locations $X \in [x_1, \cdots, x_i]$ using dataset $D$ and kernel $k(\cdot)$. This model error takes the form of the $L_2$ norm or root-mean-squared (RMS) error, seen in Eq. (1). This data $D$, defined in Eq. (2), is collected iteratively by the robot in the environment with a control policy $g^*$ that chooses a proximal location $x_{Vmax}$ that has the highest variance (or uncertainty) $V_{pred}$ in the model prediction $\hat{f}(\cdot)$.

$$J = \left\| f(X,t) - \hat{f}(X,t,D,k(\cdot)) \right\|_2 \quad (1)$$

$$D = \begin{bmatrix} t_1 & x_{k,1} & y_{k,1} \\ t_2 & x_{k,2} & y_{k,2} \\ & \vdots & \\ t_j & x_{k,j} = r_{Vmax} & y_{k,j} \\ & \vdots & \\ t_m & x_{k,m} & y_{k,m} \end{bmatrix} \quad (2)$$

Active learning algorithms are underpinned by two components: an oracle that predicts a mean and covariance function across space $\hat{f}(\cdot)$ and a policy that suggests the next location to sample $g(\cdot)$. The oracle is typically a Gaussian process due to its highly expressive capacity (lends well to characterization) and convenient uncertainty quantification in the posterior prediction (lends well to exploration), but can be represented by any model that offers a mean and covariance function as the model output shown in Eq. (3), like a probabilistic or Bayesian neural network.

$$\hat{f}(X) \sim N(\mu, V) \quad (3)$$

A Gaussian process is a probabilistic kernel method that relies on a user definition of basis kernel, most commonly the radial basis function. The basis function heavily determines the performance of the Gaussian process in generating an accurate mean and covariance function to the true underlying function, which is unknown. While Gaussian processes are mathematically elegant and conceptually simple, the kernel definition can be constraining. Neural networks offer more flexible, adaptable bases to represent a wider range of underlying functions but need more data and training time to generate an accurate model. Neural networks excel in applications of large data, complex bases, and unconstrained training time. Gaussian processes excel in applications of sparse, unevenly distributed data but can be computationally prohibitive for large datasets due to the single matrix operation that relies on matrix inversion.

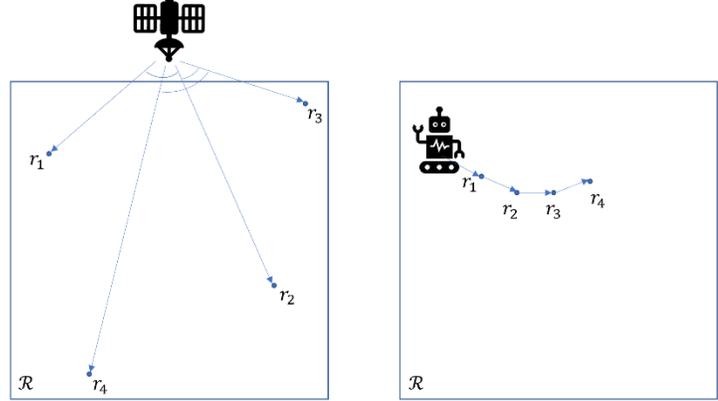

Figure 2: Difference between sampling in remote sensing (left) vs. in-situ exploration (right) applications

For the sake of exploration, a policy $g$ chooses a location $r_{Vmax}$ that has the highest variance (or uncertainty) $V_{pred}$ in the model prediction $\hat{f}$ in some space $\mathcal{R}$.

$$r_{Vmax} = g(r \in \mathcal{R}) = \underset{r \in \mathcal{R}}{\mathrm{argmax}}\ \hat{V}_{pred} \quad (4)$$

In conventional active learning algorithms, the suggestion policy is free to select the location of high uncertainty to sample across the entire global space, like a satellite leveraging remote sensing that can point any visible point on the Earth's surface depicted in Figure 1. But for applications involving in-situ sampling, like a robot visiting a destination in an environment and sampling at that specific destination depicted in Figure 2, an agent is limited to sampling at locations within finite distance. This sequence of sample locations $x_D = [x_{k,1}, \cdots, x_{k,m}]$ can be thought of as a constrained trajectory and adds nuance to how a suggestion policy may be crafted, namely defining 1) the distance between sequential samples $d_{samples}$ and 2) the uncertainty horizon to consider sampling $d_{horizon}$. Given in Eq. (5), the constrained trajectory suggestion policy is a modified version of the aforementioned unconstrained suggestion policy.

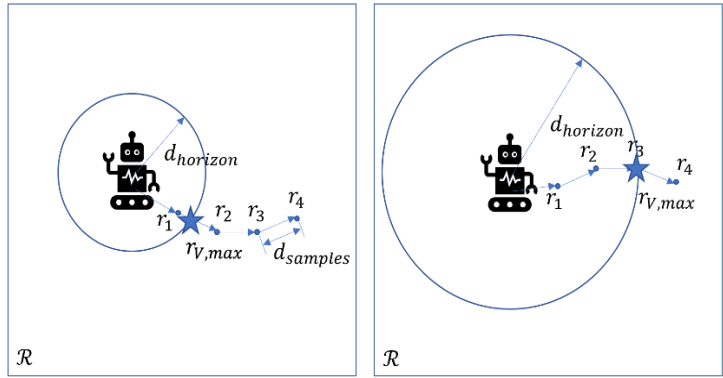

Figure 1: Difference between prediction horizon $d_{horizion}$ and sampling distance $d_{samples}$

$$r_{Vmax} = g(r \in d_{horizon}) = \underset{r \in d_{horizon}}{\mathrm{argmax}} \; \hat{V}_{pred} \qquad (5)$$

While Gaussian processes dominate in active learning literature [7]–[9], Bayesian neural networks are increasingly popular models to underpin active learning strategies, particularly for exploration of image and language tasks [10]. Neural networks and Gaussian processes are theoretically identical in the limit as a single hidden layer neural network approaches an infinite number of neurons [11]. To this extent, Tsymbalov approximated Bayesian neural networks with Gaussian processes to mitigate a neural network's overconfidence in posterior estimates [12]. No paper directly compares the performance of an active learning strategy driven with a Gaussian process as opposed to a Bayesian neural network, particularly in a spatial regression application and with constrained exploration trajectories. This work builds off of Akemoto and Zhu's previous work in applying active learning strategies to planetary surfaces but expands the analysis of exploration strategies to include Bayesian neural networks as the oracle and suggestion policy [13].

## III. Methodology

### A. Experiment Design

This study seeks to address two key research questions. Firstly, this paper aims to ascertain whether active learning strategies can lead to the convergence of more accurate models with a reduced requirement for samples and shorter traversal distances in comparison to conventional science-blind methods. Secondly, this study seeks to assess the advantages and disadvantages of leveraging Gaussian processes or Bayesian neural networks as oracle models and suggestion policies in active learning strategies.

To test these hypotheses, we characterize the performance of two exploration configurations that range in intelligence and movement constraints. Science blind exploration strategies, like a spiral or snake trajectory, hold no intelligence and offer a baseline of performance, an expected minimum outcome in test. The pre-defined science-blind trajectories offer a sequence of training data to Bayesian neural networks (BNN) and Gaussian processes (GP), which generate predictions about the target output across test inputs. These baseline results are systematically compared with results generated by experiments run across various surfaces and algorithm hyperparameters using active learning techniques in which BNNs and GPs each act as both the oracle model and suggestion policy. Active learning strategies with motion constraints embody the case study of an intelligent rover exploring an environment in situ, which is the emphasis of this paper's contribution. These active-learning-based exploration strategies driven by Bayesian neural networks and Gaussian processes are compared across diverse surface types and algorithm hyperparameters, with a focus on performance metrics such as accuracy, training time, and the number of samples required for convergence. This comparison between intelligence and motion constraint illustrates the strengths and weaknesses of the proposed constrained active learning strategies driven by Bayesian neural networks and Gaussian processes.

### B. Model Selection

To facilitate a simulation-based comparison of active learning strategies, the Bayesian neural network was built with the deepxde python package [9] and the Gaussian process model was built with the GPyTorch package [10]. Tensorflow1 [13] serves as the backend for executing the neural network implementation. The Bayesian neural network consists of 3 hidden layers, 50 neurons per layer, with sigmoid activation functions and is trained for 10,000 epochs every time a sample is added to the training set. The neural network parameters were randomly initialized with Glorot uniformity. The loss function includes an $L_2$ regularization weight of 1e-5 and the network regularizes parameters with a dropout rate of 0.01. The Gaussian process model adopts a radial basis function (RBF) kernel with the kernel's length scale optimized through gradient descent over 100 iterations. The code utilized in this study can be found in the following repository: https://github.com/xfyna/AL-BNN-GP.git.

### C. Experiment Procedure

Each experiment follows the same procedure, outlined below:

1. Load the environment's geometry (parabola, Townsend, or lunar crater), size (length and width of surface), and noise level (random Gaussian noise ranging in variance). Selection of these environment geometries are discussed in the following section.
2. Define the exploration strategy (spiral, snake, or active learning) and stopping condition (number of total samples in the training dataset divided by two for active learning strategies, detailed for each surface: parabola – 219 samples, Townsend – 219 samples, 3km Lunar – 83 samples, 6km Lunar – 311 samples).
3. Define the Gaussian process model and the Bayesian neural network hyperparameters as defined in Model Selection.
4. Initialize the agent's starting location.
5. Seed the training dataset with 10 training points.
    a. For spiral and snake methods, predefined pairs are utilized throughout the entire experiment.
    b. For active learning methods, a random walk generates the initial training data.
6. Explore the surface $R$ until a predefined maximum number of samples is reached.
    a. For spiral and snake methods, continue to sample and train the model along the predefined trajectory.
    b. For active learning:
    c. Train the Gaussian process and Bayesian neural network models on the $n$ input-output pairs in the training set thus far $(X, Y) \rightarrow \hat{f}$ where $X = \begin{bmatrix} \vec{x}_1 \\ \vdots \\ \vec{x}_n \end{bmatrix}$ and $Y = \begin{bmatrix} y_1 \\ \vdots \\ y_n \end{bmatrix}$.
    d. Predict scalar expected values $\hat{Y}_{pred}$ and variance $\hat{V}_{pred}$ in the prediction horizon $r_{pred}$.
    e. Generate a control policy $g^*$ that identifies the location in the prediction horizon with the highest variance $r_{Vmax}$ defined in Eq. (6).

$$r_{Vmax} = g^*(r \in r_{pred}) = \underset{r \in r_{pred}}{\operatorname{argmax}} \hat{V}_{pred} \tag{6}$$

    f. Traverse to the nearest neighbor location in the direction of the high-variance location $r_{Vmax}$. The action $a$ is the next location, given in Eq. (7).

$$a = \underset{r_{way} \in d_{samples}}{\operatorname{argmin}} \left\| r_{way} - r_{Vmax} \right\|_2 \tag{7}$$

    g. Sample the value $y_{n+1}$ at this location $a = x_{n+1}$ and append this training pair to the training set.

**D. Benchmark Surfaces**

Each algorithm's performance is evaluated through its ability to map an environment across three distinct surfaces: parabola, Townsend, and the lunar south pole crater. Due to the surfaces' changing complexities, the Bayesian neural network and Gaussian process machine learning strategies can be evaluated on their performance in these varying conditions, allowing a deeper understanding of their strengths and weaknesses in relation to the complexity of the environment. The inclusion of multiple surfaces facilitates a comprehensive assessment of algorithm adaptability across diverse environments, thus enhancing its robustness. Moreover, utilization of the lunar surface enables the testing of these frameworks in a real-world setting.

These surfaces have two independent dimensions (planar position $r = (x_1, x_2)$) and a third dependent dimension $y$; the dependent variable's algebraic relationship to position is known for the parabola and Townsend benchmark surfaces but unknown for the lunar ice data. The parabola surface is defined by Eq. (8), where $\sigma_{noise}^2 = 0.02$ or 0, $x_1 \in [-1: 0.1: 1]$, and $x_2 \in [-1: 0.1: 1]$. The Townsend surface is defined by Eq. (9), where $\sigma_{noise}^2 = 0.02$ or 0, $x_1 \in [-2.5: 0.1: 2.5]$, and $x_2 \in [-2.5: 0.1: 2.5]$.

$$y = x_1^2 + x_2^2 + \sigma_{noise}^2 \tag{8}$$

$$y = -\left(\cos\left((x_1 - 0.1)x_2\right)\right)^2 - x_1 \sin(3x_1 + x_2) + \sigma_{noise}^2 \tag{9}$$

The lunar surface, derived from LAMP data [14], consists of a digital elevation map (DEM) $(r = (x_1, x_2, x_3))$ of the lunar south pole in 5 m spatial resolution and hydroxyl data $y$ in 250 m spatial resolution. Noise is present in the data and significant gaps appear near the crater rim. The results comparing the six exploration strategies is presented in order of ascending complexity of the surfaces shown in Figure 3: noiseless parabola, noisy parabola, noiseless Townsend, noisy Townsend, 3km lunar crater swath, and 6km lunar crater swath.

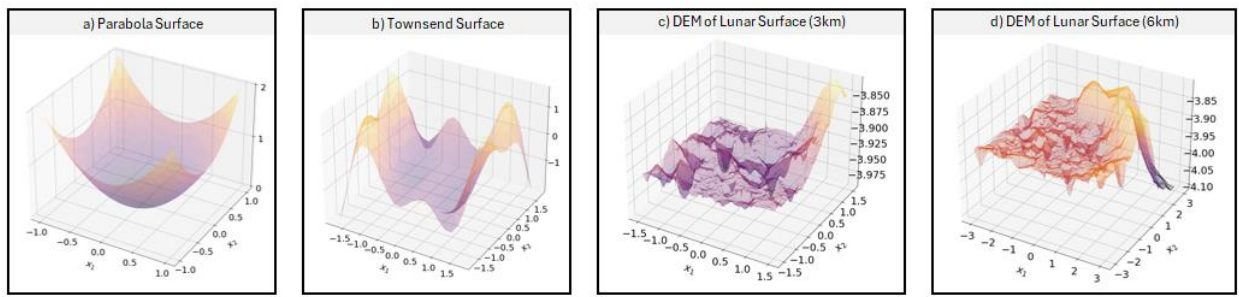

**Figure 3: Surface environments that the agent traverses: a) parabola surface, b) Townsend surface, c) 3km edge lunar crater elevation, and d) 6km edge lunar crater elevation.**

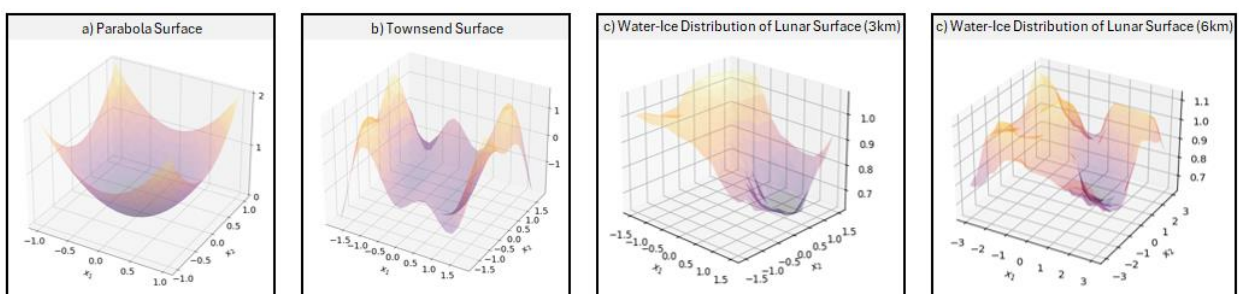

**Figure 4: True value of target outputs the rover learns: a) parabola surface elevation, b) Townsend surface elevation, c) LAMP data across the digital elevation map for the 3km lunar crater swath, and d) LAMP data across the digital elevation map for the 6km lunar crater swath**

### E.  Simulation Experiment Campaign

The Gaussian process and Bayesian neural network informed exploration strategies of various movement and prediction horizons (Table 2) were evaluated on the three surfaces of varying surface size, toggling between a noiseless and noisy measurements (Table 1). The movement horizon is varied between one grid space (movement to nearest neighbor) for active learning methods and two to four grid spaces for the science-blind snake and spiral exploration strategies. For the active learning strategies, the prediction horizon is set to look at one grid space ($1\Delta r$), three grid spaces ($3\Delta r$), or globally across the entirety of the surface ($r \in R$). By varying these parameters, the exploration efficiency of each model can be analyzed thoroughly. Although different movement horizons are compared between the science-blind and active learning methods, the science-blind method serves as a baseline metric over a pre-determined path, unlike the active learning strategy that changes with each "step" the agent takes. Additionally, note these tests are run for multiple trials to verify the validity of the data.

**Table 1: Range of simulated environments**

| Environment Topography | Size of Surface | Noise Level |
|---|---|---|
| Parabola | $x_1 \in [-1: 0.1: 1]$ | Noiseless |
| | $x_2 \in [-1: 0.1: 1]$ | Noisy |
| Townsend | $x_1 \in [-1.75: 0.1: 1.75]$ | Noiseless |
| | $x_2 \in [-1.75: 0.1: 1.75]$ | Noisy |
| Lunar Crater LAMP data | $x_1 \in [-1.5: 0.25: 1.5]$ $x_2 \in [-1.5: 0.25: 1.5]$ | Noisy |
| | $x_1 \in [-3: 0.25: 3]$ $x_2 \in [-3: 0.25: 3]$ | Noisy |

**Table 2: Range of exploration strategies and number of trials represented in results**

| Exploration Strategy | $d_{sample}$ | $d_{horizon}$ | # of Trials Each |
|---|---|---|---|
| Bayesian Neural Network Spiral | $2\Delta r$ (parabola/lunar) | | 3 |
| | $4\Delta r$ (Townsend) | | 3 |
| Gaussian Process Spiral | $2\Delta r$ (parabola/lunar) | | 3 |
| | $4\Delta r$ (Townsend) | | 3 |
| Bayesian Neural Network Snake | $2\Delta r$ (parabola/lunar) | | 3 |
| | $4\Delta r$ (Townsend) | | 3 |
| Gaussian Process Snake | $2\Delta r$ (parabola/lunar) | | 3 |
| | $4\Delta r$ (Townsend) | | 3 |
| Gaussian Process Active Learner (GPAL) | $1\Delta r$ | $1\Delta r$ (nearest neighbor) | 3 |
| | $1\Delta r$ | $3\Delta r$ (local) | 3 |
| | $1\Delta r$ | $r \in R$ (global) | 3 |
| Bayesian Neural Network Active Learner (BNNAL) | $1\Delta r$ | $1\Delta r$ (nearest neighbor) | 3 |
| | $1\Delta r$ | $3\Delta r$ (local) | 3 |
| | $1\Delta r$ | $r \in R$ (global) | 3 |

## IV. Results

### A. Metrics

To comprehensively assess the performance of Gaussian process and Bayesian neural network active learning exploration strategies, the following metrics were utilized with their associated definitions and desired intention.

- Training Time
  - All experiments were run on the same compute node on University of Hawaii's high-performance computing cluster that leveraged 32 CPUs and 128 GB of RAM. Execution time was measured using the operating system's time function prior to and after either model's training function call. Training time is a proxy for the computational intensity of either implemented algorithm.
- RMS Error upon Convergence $e_c$
  - To determine the concept of convergence loosely, the 2% settling time from control theory was adopted. The global RMS error between the model prediction and true values is inspected to verify that 1) there are enough data points to confirm convergence and 2) that the final values of RMS error stay within a 2% band of the final value $e_f$. The 2% error band $e_{2\%}$ is found by differencing the initial RMS error $e_0$ and final RMS error $e_f$, given in Eq. (10). The RMS error upon convergence $e_c$ is thus defined as the upper bound of this error band, defined by Eq. (11).

$$\Delta e_{2\%} = 0.02(e_0 - e_f) \tag{10}$$
$$e_c = e_f + e_{2\%} \tag{11}$$

It is important to note that RMS error upon convergence and along with samples/distance until convergence (mentioned in detail below) are particularly relevant for trials that exhibit asymptotic behavior, implying a convergence to a constant value. However, such convergence can only be speculated when the terminal value is unknown.

- Samples until Convergence $i_c$
  - The index of convergence or samples until convergence $i_c$ is then found through minimizing the difference between the error at an index $i$, $e_i$, and the error upon convergence $e_c$, given in Eq. (12).

$$i_c = \underset{i}{\mathrm{argmin}} \|e_i - e_c\|_2 \tag{12}$$

- Distance until Convergence $d_c$
  - The distance traveled until convergence is the sum of radial difference between each waypoint until the sample of convergence $i_c$, given in Eq. (13).

$$d_c = \sum_{i=1}^{i_c} \|x_{k,i+1} - x_{k,i}\|_2 \tag{13}$$

Note that distance until convergence can provide insight regarding which methods are more effective, as lower distance traversed until convergence implies a more efficient exploration strategy

- Position Error in Identifying Location of Global Minimum $e_{\min}$

- Eq. (14) calculates the difference between the location of the true minima and the minimum converged upon by the exploration algorithm, where the true location of the minima of the target surface for the parabola is $r_{min} = (0, 0)$, the Townsend $r_{min} = (-1.75, -1.75)$, and the lunar surface $r_{min} = (1, 0.5)$.

$$e_{\min} = \left\| r_{min} - \underset{r}{\operatorname{argmin}} \hat{f}(r \in \mathcal{R}) \right\|_2 \tag{14}$$

## B. Resulting Simulations

The variety of simulations aims to emphasize the difference in performance between the exploration strategies embedded with differing models and to specifically highlight the performance of the constrained active learner. The science-blind algorithms, snake and spiral alike, offer baseline performance metrics to compare the effectiveness of intelligent strategies. The constrained active learning algorithms aim to mimic rovers, the main interest of this paper. Simulations of exploration using the constrained active learner, illustrated in Figures 5 to 7, display specific exploration trajectories and model performance over the iterative sampling experiment.

As previously displayed in Table 2, there are three different prediction horizons associated with the algorithms that utilize active learning and a constrained movement horizon: nearest neighbor (NN), local, and global prediction horizons. Each simulation of an exploration algorithm generates figures to illustrate the evolution of the underlying model's performance. Figures 5 - 7 are formatted such that the top three graphs display data regarding the BNN algorithm, the middle three graphs display data regarding the GP algorithm, and the bottom two graphs compare the GP and BNN performance, with GPs being graphed in blue and BNNs graphed in black. Figure 5 has each subplot labeled. Subplot a) illustrates the BNN prediction across the surface test location, where the colored surface is the ground truth and the gray surface is the prediction. The purple star represents the agent's location on the surface and the black lines represent the agent's historical trajectory. Subplot b) displays the BNN algorithm's uncertainty across the environment at the model's most recent evaluation. Next, subplot c) displays the BNN algorithm's error across the surface at the most recent model evaluation. Subplots d), e), and f) show similar information as the plots above them, but for a GP algorithm. Lastly, subplot g) graphs the RMS error and subplot h) graphs the variance, which is defined as the mean of the uncertainty graphed in plots b) and e).

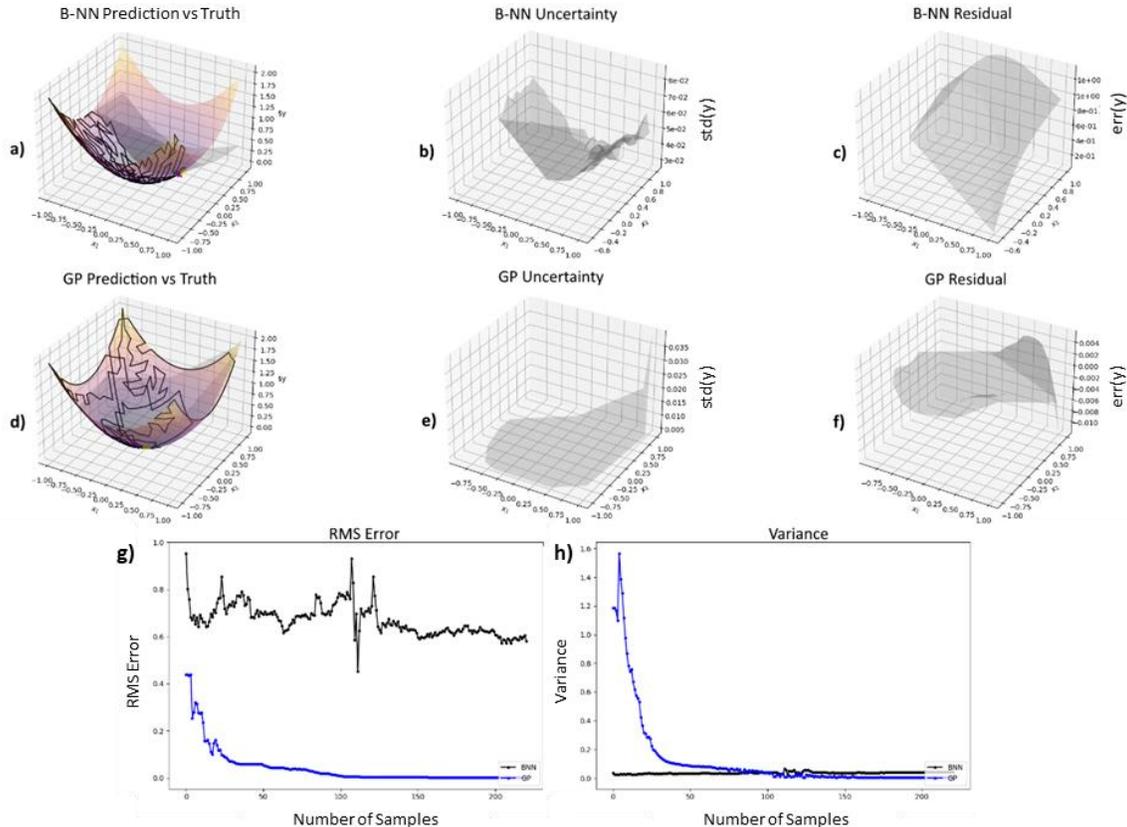

**Figure 5: Example of an active learning GP and BNN model with a local prediction horizon and constrained movement horizon on a noiseless parabola**

Figure 5 illustrates a constrained active learning algorithm comparison between a BNN and GP with a local prediction horizon across a noiseless parabola. The GP active learning strategy initially preferences the agent to traverse the outer edges of the surface. The agent then moves inward and explores the remainder of the surface in a set number of total samples. The BNN active learning strategy explores one half of the space thoroughly. These behaviors carry onto the constrained AL comparison but with a nearest neighbor prediction horizon across a Townsend surface, shown in Figure 8. GP driven active learning with nearest neighbor and local prediction horizons demonstrated the best overall performance across all active learning strategies. The results derived from trials utilizing these algorithms were not only consistent across trials, but also the most precise in finding the global minimum and on the higher end of computational efficiency.

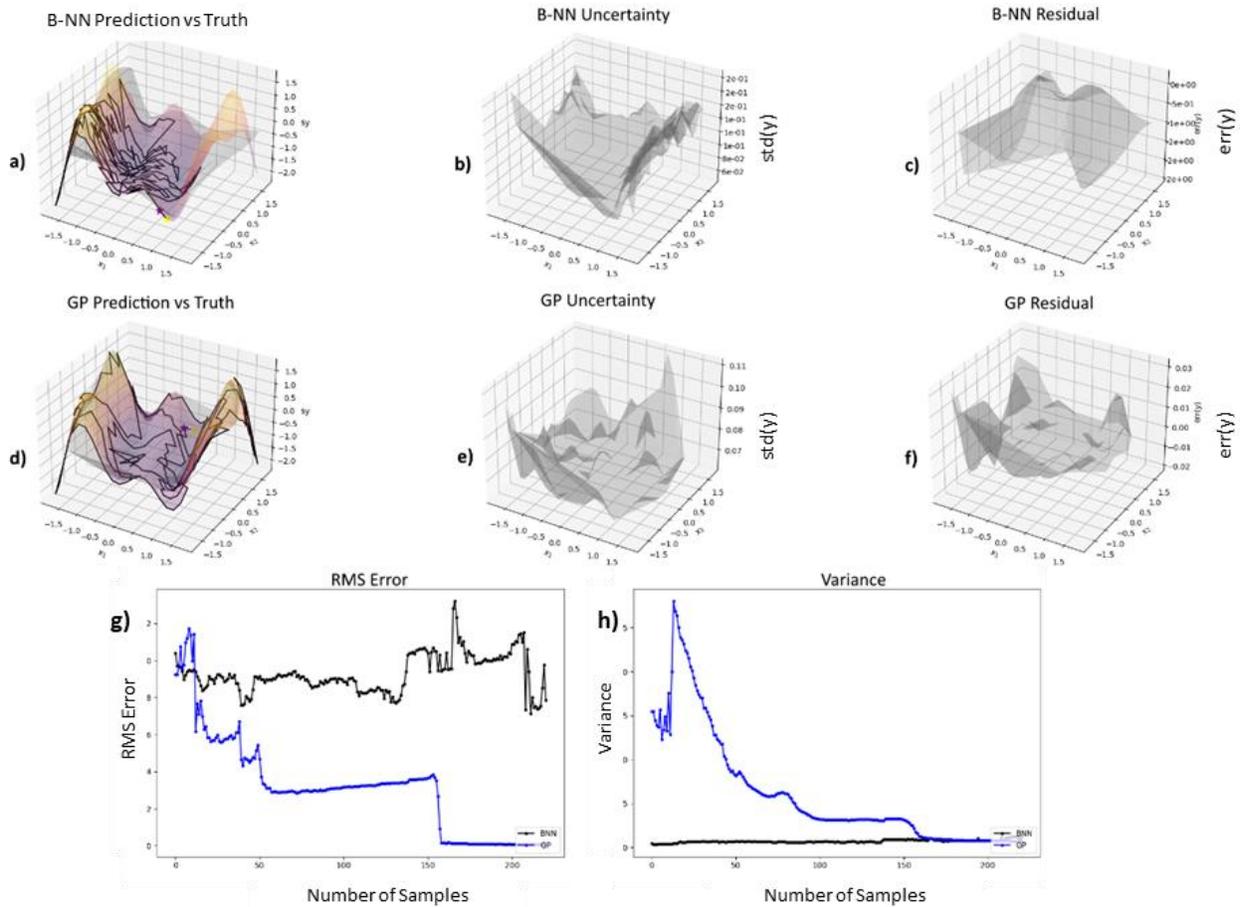

**Figure 6: Example of an active learning GP and BNN model with a nearest neighbor prediction horizon and constrained movement horizon on a noiseless Townsend**

Figure 7 displays algorithm performance when a global prediction horizon is imposed on a constrained trajectory active learner. Although still one of the higher performing algorithms, the GP driven global prediction horizon algorithm does not compare to the efficiency reached with GP algorithms of smaller prediction horizons. Instead of traveling across the edges, the agent mimics the BNN algorithm's movement patterns of oversampling a region of the space. Consequently, the GP algorithm ends with higher error in finding the global minimum, as compared to other exploration strategies that utilize GPs. Along with this, the GP algorithm requires increased samples and distance to reach convergence.

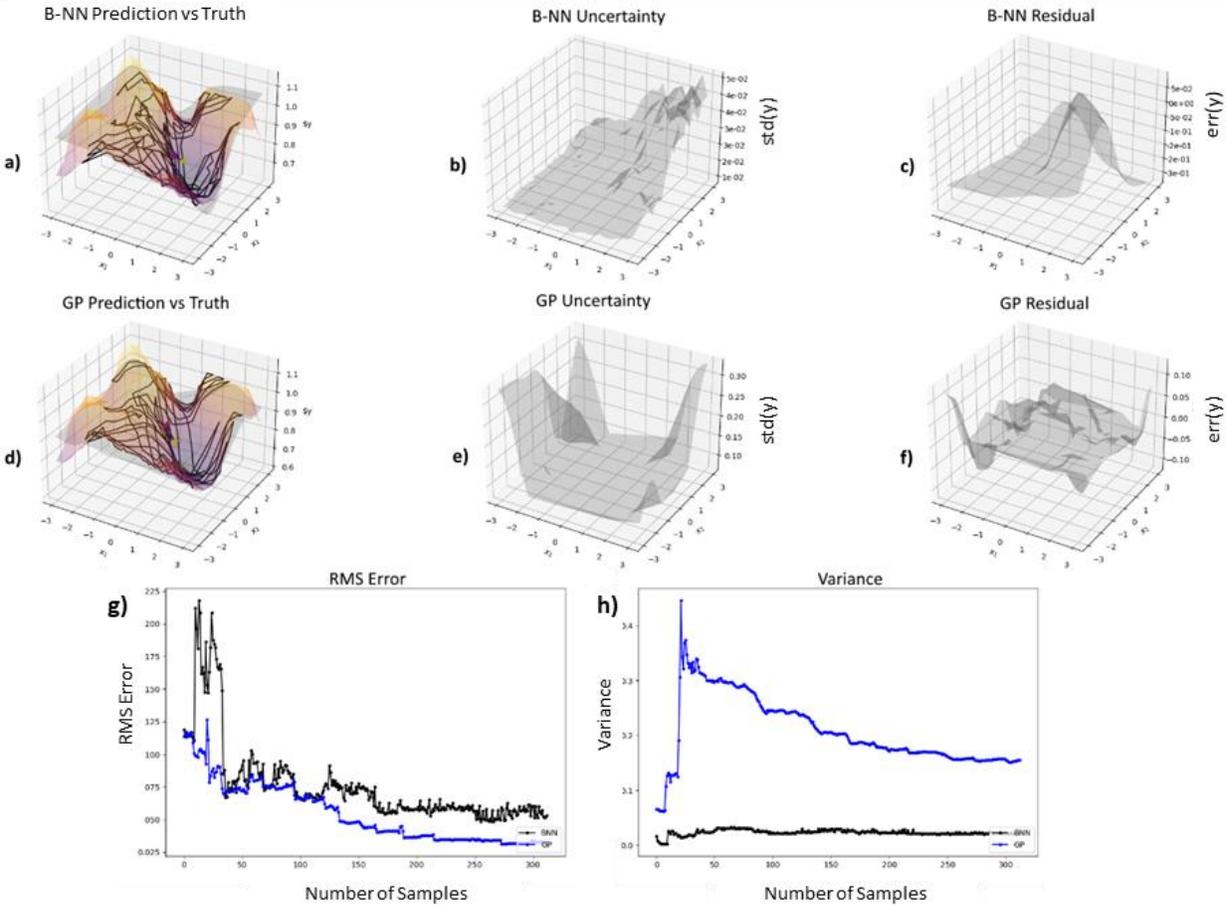

**Figure 7: Example of an active learning GP and BNN model with a global prediction horizon and constrained movement horizon on a 6km lunar crater swath**

### C. Analysis

Across all surface environments and algorithm hyperparameters, the Gaussian process active learner required less time to train the model for training datasets containing up to 331 samples, the size of the largest environment GP active learning algorithms are generally more accurate than the BNN active learning algorithms with some exceptions. GPs usually converge to a good model in less samples than a BNN. Active learners (BNN and GP alike) require less roving distance to converge to an accurate model, though not much less roving distance than science blind methods. GPs are more accurate in identifying the surface's true minimum location. These findings underscore the distinct advantages of GP-based active learning strategies in optimizing training efficiency and accuracy across diverse surface environments for constrained movement horizons.

The results are displayed in the order of the metrics defined in Section A. For nearly all plots, the performance metrics were plotted in a log scale across the y-axis for all figures, except when analyzing the error in locating the global minimum. The x-axis spans the surface type and exploration strategy. Exploration strategy is defined by the acronyms "SB" and "AL", which are acronyms respectively for science-blind and active learning. The movement horizon is indicated after the exploration strategy definition and is categorized as either $1\Delta x$, $2\Delta x$, or $4\Delta x$. Each point in the figure represents a mean value and the error bars around each point represent standard deviation across all noiseless/noisy trials completed for each exploration strategy. Note that results for both snake and spiral science-blind strategies are captured into one data point. Lastly, whether the algorithms are driven by GPs or BNNs is denoted on the legend with the use of "GP" or "BNN", as well as color coordination.

*Training Time*

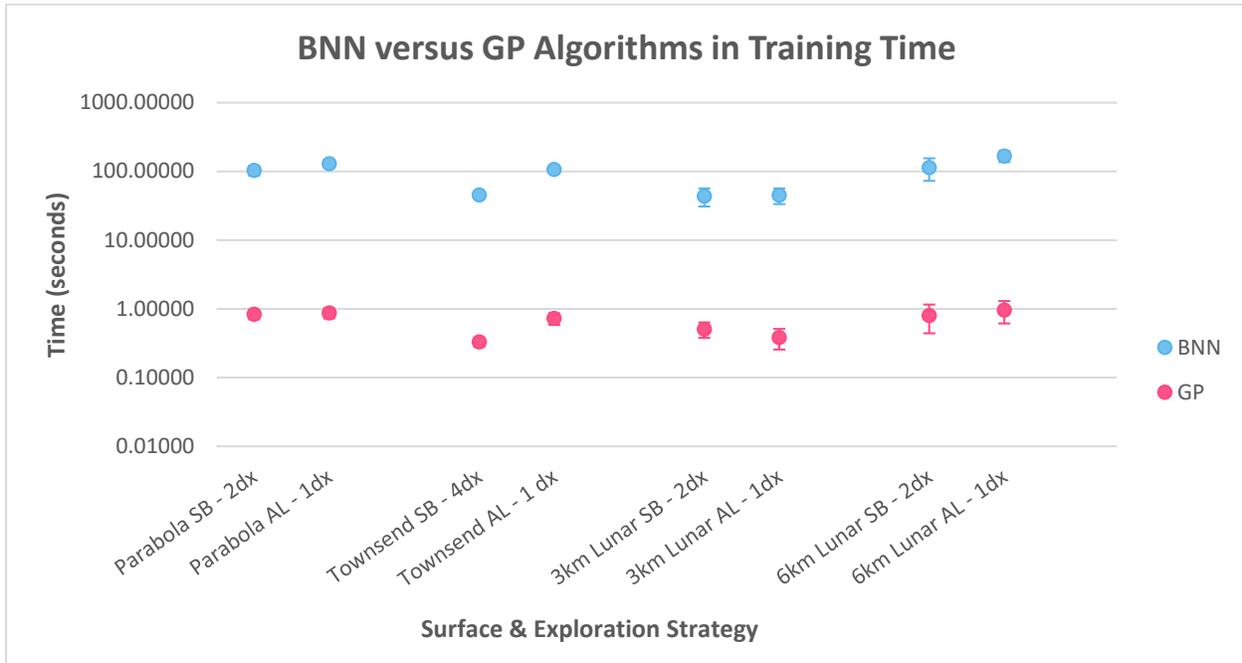

**Figure 8: Comparison of BNN and GP science blind and active learning strategies on training time**

GP algorithms have a shorter training time compared to their BNN counterparts across all surface types as seen Figure 8 below. The training time across science-blind and constrained active learning strategies did not differ much when comparing GP or BNN algorithms individually. GP algorithms are generally more computationally efficient than BNN algorithms.

Figure 9 highlights a single trial in which training time per each sample is graphed. There appears to be a small increase in training time for the BNN algorithm as the number of samples increases. For example, this BNN algorithm started at 23.3 seconds per sample and ended with 35.3 seconds per sample. This instance of a GP algorithm maintained more steadiness with 0.308 seconds per sample initially and moved to 0.288 seconds per sample by the end of the simulation.

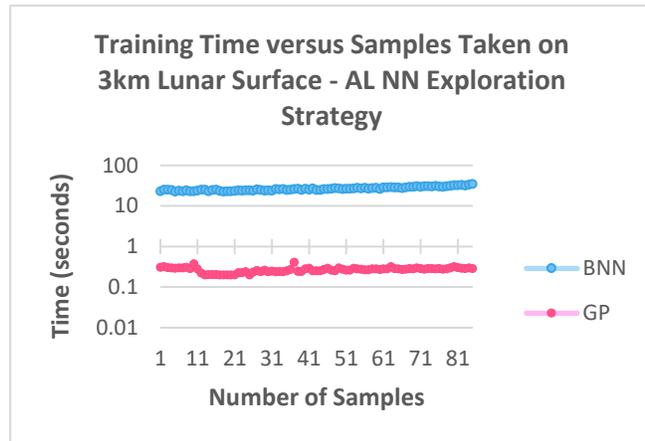

**Figure 9: Training time versus samples taken across a 3km lunar surface utilizing the constrained active learning NN exploration strategy**

*RMS Error Upon Convergence*

GP algorithms outperform BNN algorithms, showing lower average RMS error, as shown in Figure 10 below. There are a few exceptions to this trend. One deviation can be witnessed in the baseline science-blind exploration strategy across the Townsend surface, where the GP science-blind algorithms have a higher RMS error as compared to their BNN counterparts. This could be due to the complexity of the surface, which may require a more expressive basis function to model it effectively. Of note, the science-blind snake method performed poorly on the Townsend surface and did not converge; therefore, the only data shown comes from the science-blind spiral method. Figure 10 illustrates that on average, GP algorithms produce higher state accuracy as compared to BNN algorithms. Comparing the active learning algorithms to the science-blind methods, note that as the surface complexity increases, there appears to be a increase in the RMS error for SB strategies. This can be seen through the Townsend surface, as well through the lunar crater, where the RMS error for SB methods approaches that of the active learner.

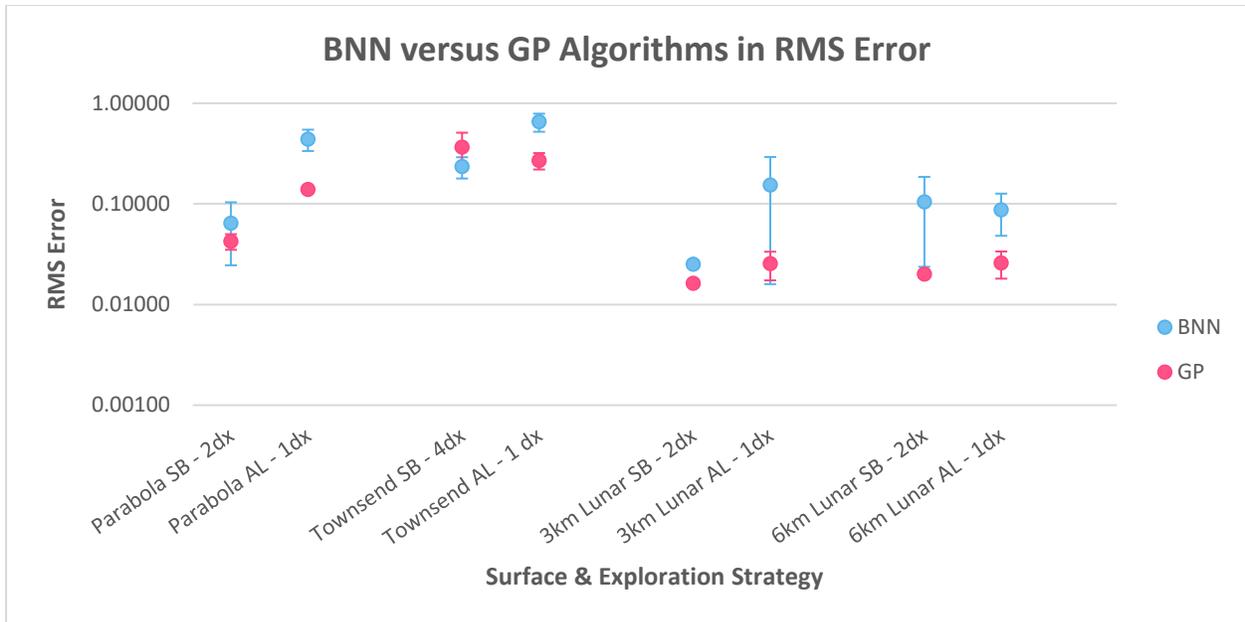

**Figure 10: Comparison of BNN and GP science blind and active learning strategies on RMS error**

*Samples Until Convergence*

    The effectiveness and superior performance of GP algorithms continues to be displayed in Figure 11, which details the samples until convergence is reached. The GP algorithms display lower sample count until convergence in all instances of exploration strategy and movement horizon, except one case. Figure 11 illustrates the instance in which the GP algorithm does not converge with lower sample numbers across the 6km lunar crater surface where the baseline science-blind GP algorithm does not outperform the BNN algorithm in terms of number of samples taken. Regardless of this baseline metric, the active-learning strategies that utilize GP algorithms take lower samples than their BNN counterparts. In regards to the science blind methods, lower samples were taken in every instance (due to the nature of the pre-determined path), and therefore there were less samples available upon evaluation of convergence. Regardless, the graph below demonstrates that science-blind methods converged to a global minimum in less samples than the active learning strategies.

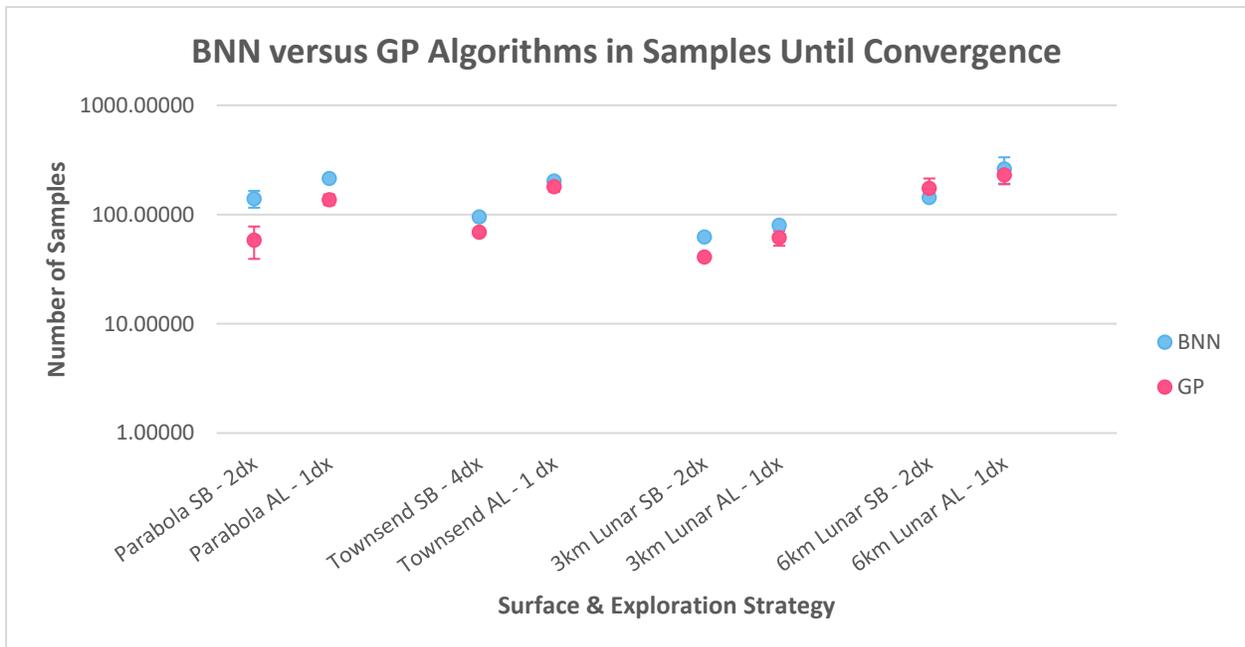

**Figure 11: Comparison of BNN and GP science blind and active learning strategies on sample until convergence**

*Distance Until Convergence*

Data regarding the distance traveled until reaching convergence is illustrated in Figure 12. BNN algorithms generally travel farther distances than GP algorithms to converge on a model, suggesting decreased effectiveness as compared to GP models. The 6km lunar surface displays an exception to this trend where the snake science-blind GP algorithm cannot converge on an accurate model over the 6km lunar crater surface. The GP algorithm requires a slightly higher distance for convergence as compared to the BNN algorithm. Again, the increased complexity of the surface may require a more expressive basis function, which is provided by the BNN algorithm in this instance. This data ultimately confirms that GP algorithms generally perform better when utilizing active learning exploration strategies over science-blind methods.

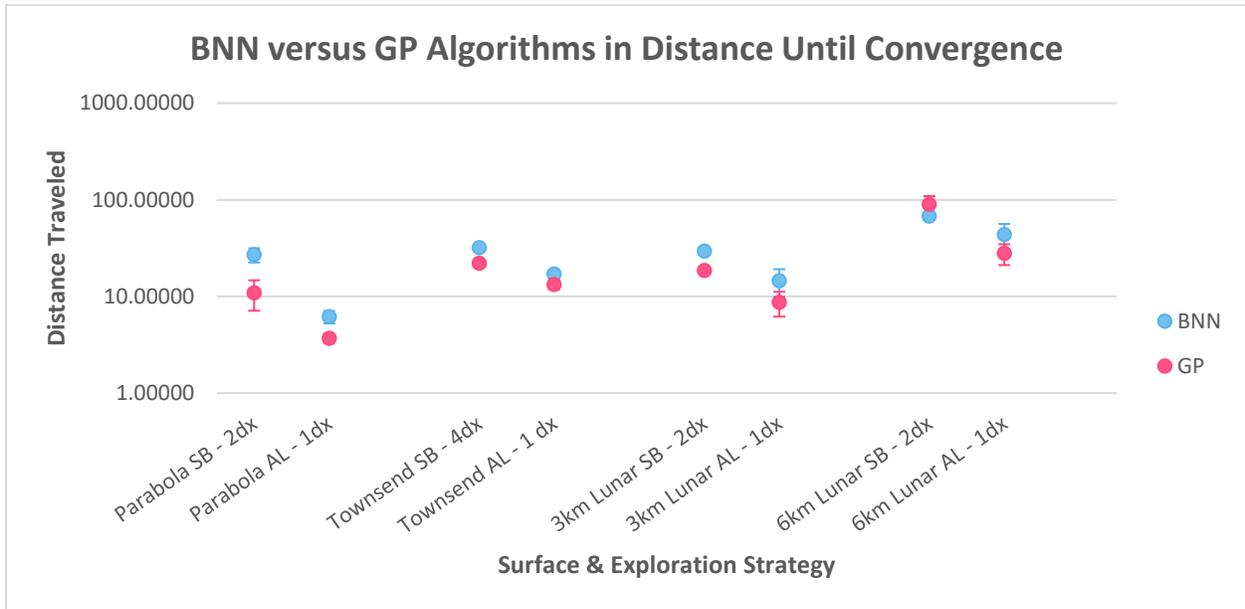

**Figure 12: Comparison of BNN and GP science blind and active learning strategies on distance until convergence**

*Position Error in Identifying Location of Global Minimum*

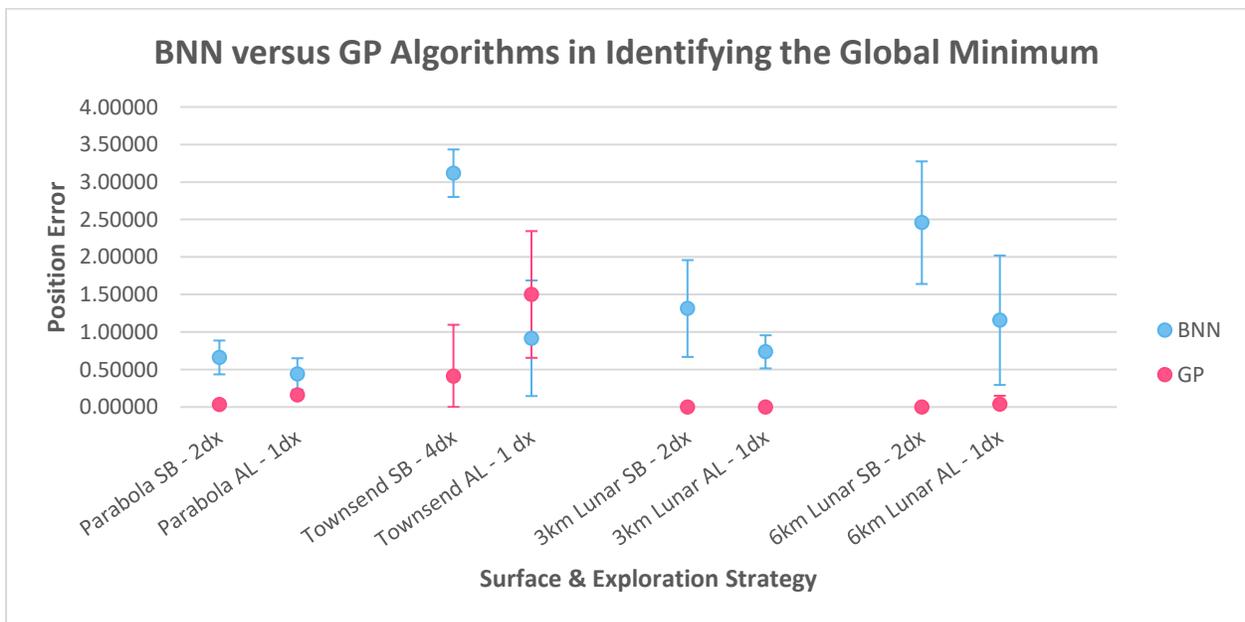

**Figure 13: Comparison of BNN and GP science blind and active learning strategies on position error in finding the global minimum**

The GP algorithms continue to outperform the BNN algorithms, as seen in Figure 13 where all active learning GP algorithms have lower average position error in finding the global minimum. In fact, the GP algorithm converges to the correct global minimum with zero position error in two instances and three other instances approach near zero error. None of the BNN models could precisely identify minima locations across any surface. In terms of the science-blind methods, although convergence to a low position error global minimum did occur, it is not suggestive of a better performing exploration strategy. This is due to the fact that the science-blind method traversed an environment in a meticulous way that required the agent to travel farther than was necessary. As such, the active learners provide a more cost-effective strategy.

## V. Conclusion

This paper investigates the comparative performance between active learning algorithms driven by Gaussian processes and Bayesian neural networks tested in various simulation environments to predict the spatial distribution of a variable of interest along multiple datasets. The active learning algorithms consistently converge to an accurate model after traversing less distance as compared to science-blind methods. Note that a smaller distance traveled does not signify fewer samples were taken, as science-blind methods require fewer samples to reach convergence than their active learning counterparts. We can also conclude that GP models are superior oracles for active learning strategies, provide higher computational efficiency, and predict with more accuracy than BNN models across all environments tested. GP algorithms outperform BNN algorithms in nearly all cases, with the exception being when target surface is very complex or when global prediction horizons are utilized. Instead, GP algorithms benefit from short-sightedness where greedy actions lead to increased rewards.

This model has potential for future applications in rovers traversing planetary surfaces, such as the Moon or Mars. Not only do these algorithms have the capability to assist in the search for water-ice on these surfaces, but can be easily extended to search for other science objectives. The authors recommend that Gaussian process oracle models assist in science operations whether in real time onboard the rover or as a suggestion system to teleoperators offline. The next step in furthering this research is to encode the GP algorithm onto a physical rover and conduct field testing with real time science data.

## Acknowledgments

This work was supported by NASA Grant HI-80NSSC21M0334. We would like to extend our sincerest appreciation to the University of Hawaii's high-performance computing (HPC) cluster IT department, who assisted in not only ensuring the full-time operation of the HPC cluster but also in solving the many technical difficulties that arose throughout the duration of our research.